\documentclass[11pt]{article} %11pt

\usepackage{graphicx}
\usepackage{booktabs}
\usepackage[T1]{fontenc}
\usepackage{lmodern}
\usepackage[flushleft]{threeparttable}
\usepackage[hidelinks]{hyperref} %
\hypersetup{hidelinks}
\usepackage{multirow}
\usepackage{longtable}
\usepackage[margin=3cm]{geometry} %
\usepackage{amsmath}
\usepackage{authblk}
\usepackage{caption}

\title{Latent space configuration for improved generalization in supervised autoencoder neural networks}
\author[1]{Nikita Gabdullin}
\affil[1]{Joint Stock "Research and production company "Kryptonite" \authorcr
E-mail: n.gabdullin@kryptonite.ru}
\date{}
\begin{document}

    \captionsetup[table]{labelformat={default},labelsep=period,name={Table}}

    \maketitle

    \begin{abstract}
        Autoencoders (AE) are simple yet powerful class of neural networks that compress data by projecting input into low-dimensional latent space (LS). Whereas LS 
        is formed according to the loss function minimization during training, its properties and topology are not controlled directly. In this paper we focus on AE LS 
        properties and propose two methods for obtaining LS with desired topology, called LS configuration. The proposed methods include loss configuration using a 
        geometric loss term that acts directly in LS, and encoder configuration. We show that the former allows to reliably obtain LS with desired configuration by 
        defining the positions and shapes of LS clusters for supervised AE (SAE). Knowing LS configuration allows to define similarity measure in LS to predict labels 
        or estimate similarity for multiple inputs without using decoders or classifiers. We also show that this leads to more stable and interpretable training. 
        We show that SAE trained for clothes texture classification using the proposed method generalizes well to unseen data from LIP, Market1501, and WildTrack 
        datasets without fine-tuning, and even allows to evaluate similarity for unseen classes. We further illustrate the advantages of pre-configured LS similarity 
        estimation with cross-dataset searches and text-based search using a text query without language models.
  
    \end{abstract}

    
    \section{Introduction}
\label{introduction}

Computer vision (CV) is one of the most important branches of computer science (CS) which encompasses image processing for image and video analysis in medicine, 
robotics, autonomous vehicles, manufacturing, and other areas. Nowadays, neural networks (NN) play dominant role as the main tool for solving CV tasks. Whereas from 
early days of Machine Learning (ML) to the era of Deep Learning (DL) more complex and larger NN model architectures have been proposed, autoencoders (AE), while being 
relatively simple, are still incredibly useful. They play major role in image processing \cite{AEenh, AEres}, recommendation systems 
\cite{Autorec}, anomaly detection \cite{AEanomaly}, and other areas of CS. 

AEs encode input as low-dimensional vector in latent space (LS) which is later transformed by decoder to obtain the output \cite{AErev}. This is commonly done in 
end-to-end fashion without any control over LS representation. However, AE classifiers and recently proposed probabilistic AEs rely strongly on the proximity of 
similar inputs in LS. In this paper we study the possibility of influencing LS directly during training and inference. We propose loss and encoder configuration 
methods which we refer to as LS configuration. We illustrate the results by visualizing 2-dimensional LS of supervised AE (SAE) trained for texture classification. 

Loss configuration is achieved by adding geometric loss term to loss function during training. We report that that pre-configured LS shows guaranteed clusterization 
pattern in space with predetermined dimensions. Training also becomes more stable and predictable with its results being more easily interpretable. Knowing LS 
configuration it is possible to define a similarity measure in LS to evaluate similarity of multiple inputs. We show that SAE trained with geometric loss generalizes 
better than the one with conventional training. For that we train SAE on a subset of Look Into Person (LIP) \cite{LIPD} dataset and use it for image similarity estimation 
and to perform query-test searches in Market1501 \cite{MarketD} and Wildtrack \cite{WildD} datasets. Moreover, we perform cross-dataset searches using Market images as 
queries and Wildtrack as test database, and vice versa. 

Finally, we show how pre-configured LS allows to correlate LS regions with texture types and use that to perform searches using text queries. Using text to prompt NN 
is not unique and is widely used in diffusion models. However, this requires an additional language model that creates text embeddings based on input text \cite{StDiff, clipp}. 
Therefore, an additional model to create embeddings that cannot be interpreted directly is required. On the contrary, we show how the direct use of LS properties can 
be used to reliably locate regions with desired textures without language models. 

The rest of the paper is organized as follows: Section~\ref{Autoencoders} provides a brief overview of AE and their types, Section~\ref{LSconfig} proposes LS configuration methods, 
Section~\ref{GLoss} formulates Geometric loss, Section~\ref{LSconfclassif} shows LS configuration experiments and results, Section~\ref{LSsim} discusses similarity in LS and 
how it can be used for image retrieval, Section~\ref{discussions} provides additional discussions, and Section~\ref{conclusions} concludes the paper.  

\begin{figure}[b]
	\centering
	\includegraphics[scale=0.45]{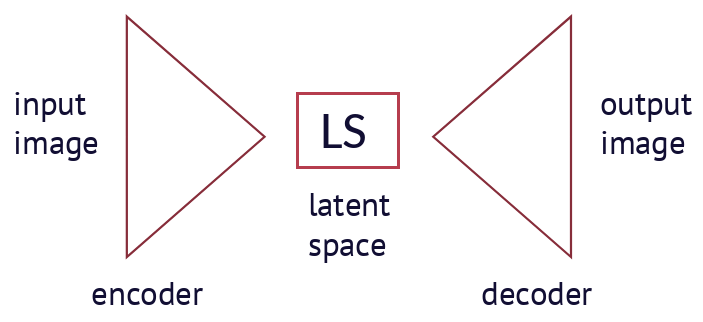} 
	\caption{Autoencoder architecture.}
	\label{fig:1}
\end{figure}
\unskip

\section{Autoencoders}
\label{Autoencoders}

\subsection{Autoencoder architecture}
\label{AE-architecture}

Autoencoders are neural networks which consist of encoder that encodes or compresses high-dimensional input data into low-dimensional LS and decoder that decodes or 
upscales LS representation for image reconstruction, as shown in Figure~\ref{fig:1}. Encoder and decoder usually consist of sequences of fully connected or 
convolutional layers. For many CV tasks AE can be trained in unsupervised manner simply by learning to reconstruct the input image. In this case Mean Square Error (MSE) 
loss is used as reconstruction loss function. This significantly simplifies training data processing and allows to use unlabeled data.

\subsection{Variational Autoencoders}
\label{VAE}

Variational autoencoders (VAE) are special probabilistic class of AEs. Encoder of VAE includes two additional layers with their outputs linked to mean \textit{$\mu$} 
and standard 
deviation \textit{$\sigma$} of the modeled distribution using (\ref{eq:vae_z}) and (\ref{eq:kld}) \cite{VAEo,VAEp}. Equation (\ref{eq:vae_z}) shows that LS 
representation \textit{z} in 
VAE contains random sampling from a normal distribution \textit{N(0,1)} with zero mean and $\sigma=1$. 
Kullback-Leibler divergence (KLD) loss is calculated in addition to MSE loss to make LS distribution more 
like \textit{N(0,1)}. VAE is trained with combined loss consisting of weighted sum of KLD and MSE. During inference VAE outputs are sampled from a distribution similar 
to the 
input which is advantages for generators. The most notable VAE is UNet \cite{UNetp} which is widely used by Stable Diffusion \cite{StDiff} and other generative models. 

\begin{equation}
	z = \mu + \sigma \cdot N(0,1),
	\label{eq:vae_z}
\end{equation}
\unskip

\begin{equation}
	L_{KLD} = \mu^2 + \sigma^2 - log(\sigma) - 0.5,
	\label{eq:kld}
\end{equation}
\unskip

\begin{equation}
	L = L_{MSE} + k_{d} \cdot L_{KLD},
	\label{eq:vae_loss}
\end{equation}

where \textit{k\textsubscript{d}} is KLD weight coefficient.

\begin{figure}[b]
	\centering
	\includegraphics[scale=0.45]{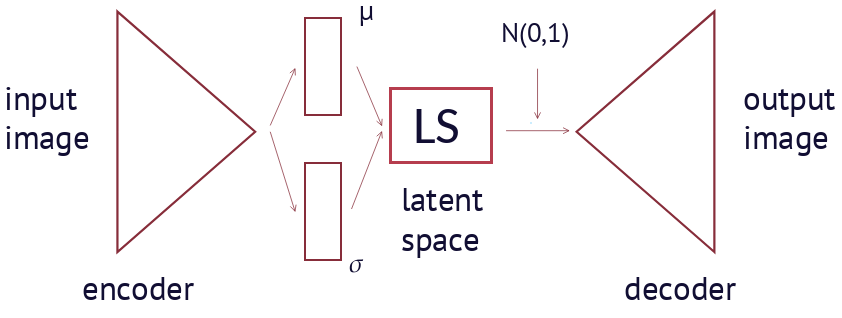} 
	\caption{VAE architecture.}
	\label{fig:2}
\end{figure}
%\unskip

\subsection{Supervised Autoencoders}
\label{SAE}

Whereas AEs were initially proposed for unsupervised learning scenarios, they were proven useful for semi-supervised and supervised learning tasks, too. As any AE, SAE 
has encoder-decoder architecture with its decoder substituted by a classifier neural network. Decoder and classifier can also be used in parallel so that decoder 
output is used for regularization \cite{SAEreg}, but this use case is outside of the scope of this study. For supervised learning the reconstruction loss is substituted 
with classification loss most commonly being Cross Entropy (CE) loss. In this paper we study LS of SAE with architecture shown in Table~\ref{tab:1} where Double layer means 
two subsequent sequences of convolution, batch normalization \cite{batchnormp}, and ReLU; and Down layer consists of MaxPool \cite{poolp} followed by Double layer. This 
encoder architecture is inspired by VAE architecture \cite{unetgit} without (\ref{eq:vae_z}) and (\ref{eq:kld}) so dense layer outputs have no statistical meaning and 
can be referred to as \textit{a} and \textit{b}, so LS representation is \textit{z} = \textit{a+exp(b)}. Classifier is a simple 3-layer fully-connected network. Inference of such SAE is deterministic due to the 
absence of \textit{N(0,1)} term in \textit{z}. 

\begin{table}
  \caption{Studied SAE architecture.} 
  \label{tab:1}
  \centering
  \begin{tabular}{|c|c|c|c|}
    \hline
    Element & Layer & Parameter & Channels \\
    \hline
    Input & - & \textit{x} & 1,32,32 \\
    \hline
    \multirow{7}{*}{Encoder} & Double(32->64) &	- & 	64,32,32    \\ \cline{2-4}    
                            & Down(64-128) &	- &	128,16,16 \\ \cline{2-4}
                            & Down(128-256) &	- &	256, 8, 8 \\ \cline{2-4}
                            & Down(256-512) &	- &	512, 4, 4 \\ \cline{2-4}
                            & Down(512-512) &	- &	512, 2, 2 \\ \cline{2-4}
                            & flatten &	- &	2048 \\ \cline{2-4}
                            & Lin(2048,2) & \textit{a} & 2 \\
                            & Lin(2048,2) & \textit{b} & 2 \\ 
    \hline
    LS & \textit{a} + \textit{exp(b)} & \textit{z} & 2 \\
    \hline
    \multirow{3}{*}{Classifier} & Lin(2,32) &	- &	32    \\ \cline{2-4}    
                            & Lin(32,64) &	- &	64  \\ \cline{2-4}
                            & Lin(64,5) &	\textit{p} &	5 \\ \cline{2-4}
    \hline
    Output &	argmax(\textit{p}) &	\textit{y} & 1 \\
    \hline
  \end{tabular}
\end{table}

\begin{figure}[b]
	\centering
	\includegraphics[scale=0.45]{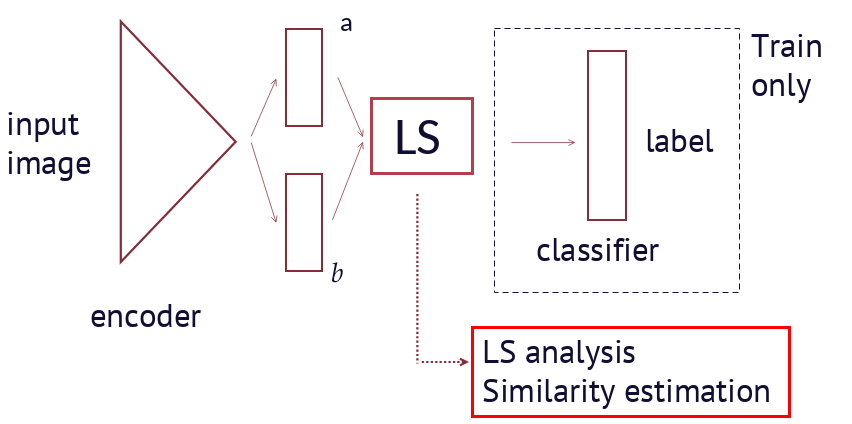} 
	\caption{Conventional encoder-classifier SAE architecture to convert LS 
  representation to label and the proposed direct LS analysis.}
	\label{fig:3}
\end{figure}
%\unskip

\subsection{SAE latent space analysis}
\label{SAE-LS}

To predict label for a given input SAE first encodes it in LS and then uses classifier to produce the label from LS representation. However, LS representation, 
containing all the necessary information to predict the label, cannot be interpreted directly using conventional methods. To address this, it was previously proposed 
to use stochastic iterative algorithms to evaluate LS structure dynamically during training and possibly avoid using classifiers \cite{SAEcentroids, SAEpca, bnneck}. 
Our approach is similar to center loss \cite{centloss} training with deterministic clusters and modified contributions of out-of-cluster samples to the loss function.
In this paper we show 
that analyzing LS representation is possible when LS topology is known, and discuss means of obtaining the desired topology. We show in Section~\ref{LSsim} that working with 
pre-configured LS allows to perform operations not readily available for conventional classifiers, such as similarity estimation for multiple inputs by analyzing their 
LS representations.

\section{Latent space configuration methods}
\label{LSconfig}

In this paper we propose two methods to configure LS in AE as shown in Figure~\ref{fig:4} which are loss configuration and encoder configuration. 
The former can be considered as a special training procedure that allows to obtain LS with desired properties. The latter is more invasive since it changes the 
model in the same way adding $\mu$ and $\sigma$ layers does to AE to obtain VAE does.

\subsection{Loss configuration}
\label{Lossconf}

\begin{figure}[h]
	\centering
	\includegraphics[scale=0.55]{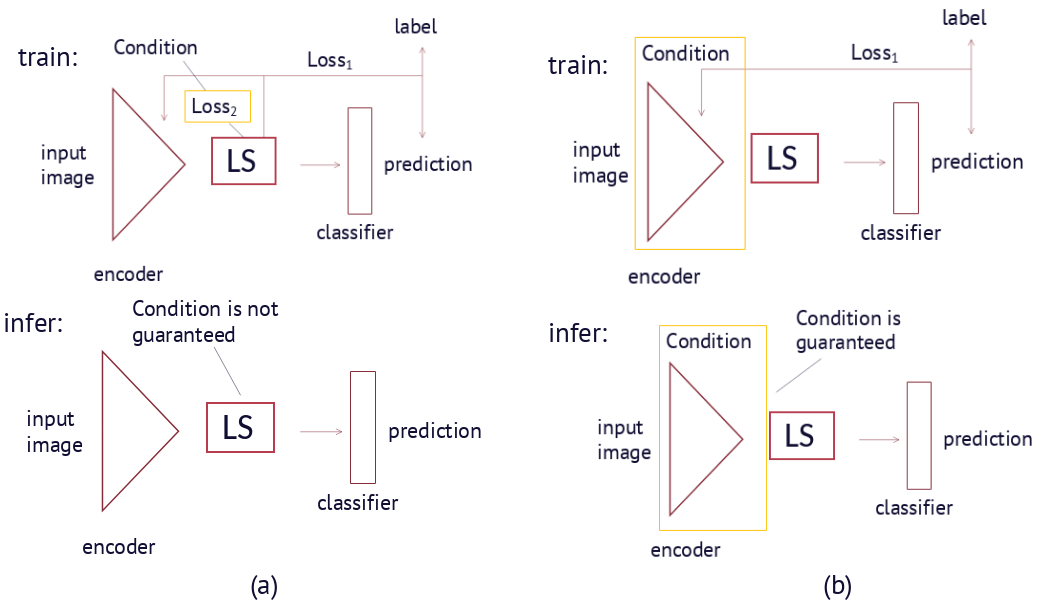} 
	\caption{Differences between loss and encoder configuration.}
	\label{fig:4}
\end{figure}

Figure~\ref{fig:4} (a) illustrates that in loss configuration scenario a loss function called geometric loss (\textit{L\textsubscript{G}}) operates directly on LS and not on classifier output.
Training is done with a loss function combining configuration loss with conventional CE or MSE loss, similar to KLD combined loss discussed in Section~\ref{VAE}. 
This method does not require modifying the model, which can be an advantage when large pre-trained models are used and their fine-tuning is desired. Loss configuration 
is discussed in detail in Section~\ref{GLoss} with experiment results presented in Section~\ref{LSconfclassif}. The main drawback of this method is that loss condition 
is guaranteed only 
during training since loss function does not influence inference directly. This can be a problem for small datasets where training data does not contain the complete 
variability of possible inputs. 

%\unskip

\subsection{Encoder configuration}
\label{Encconf}

Figure~\ref{fig:4} (b) illustrates that the main advantage of encoder configuration is that it is guaranteed for inference. However, this requires implementing the 
condition as part of the encoder, which requires to modify the model. An example of such modification is the transition from AE to VAE discussed in Section~\ref{Autoencoders}. 
This method allows more control over the LS properties than loss configuration. For instance, polar coordinates or other transformations that completely change LS 
topology can be added. To illustrate this concept, SAE that uses polar coordinates in its encoder is studied. We aim to obtain LS with sectors each of which 
accommodates a specific class. We also add prohibited LS sectors such that if input is projected into the prohibited area, its LS position is modified to the next 
allowed sector by adding an extra rotation. For instance, for the case with 6 classes we define 50 degree class sectors with 10 degree prohibited sectors and an extra 
rotation of 30 degrees for points projected into prohibited regions. To achieve that an encoder of SAE in Table~\ref{tab:1} is modified as

\begin{equation}
	r = hypot(z[0],z[1]),
	\label{eq:pol-r}
\end{equation}

\begin{equation}
	\phi = atan2(z[1],z[0]),
	\label{eq:pol-phi}
\end{equation}

\begin{equation}
	\phi' = 
	\begin{cases}
	\phi + \pi/6\ if\ (\phi\ mod\ \pi/3)\ mod\ 2 \pi/9 > \pi/18,  \\
	\phi\ otherwise;
	\label{eq:pol-phi'}
	\end{cases}
\end{equation}

\begin{equation}
	z_{new} = concat(r \cdot cos(\phi'),r \cdot sin(\phi')).
	\label{eq:z_new}
\end{equation}

\begin{figure}[h]
	\centering
	\includegraphics[scale=0.5]{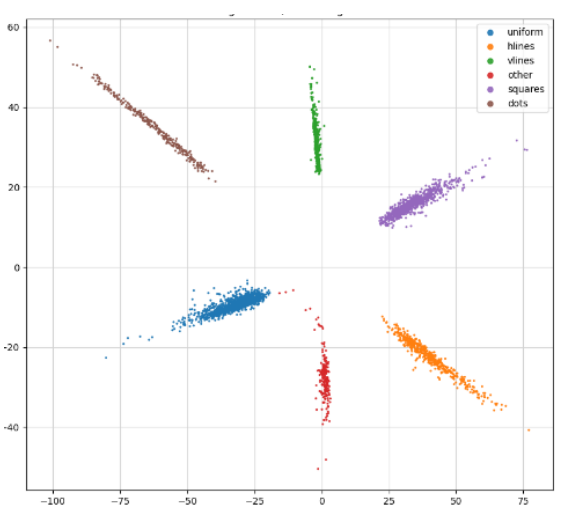} 
	\caption{SAE trained with polar coordinate condition in encoder.}
	\label{fig:5}
\end{figure}

It should be stressed that since the area prohibition is implemented in the encoder, no point can lie in prohibited areas of LS even during inference. For SAE described 
in Section~\ref{VAE} with modified encoder trained with \textit{L\textsubscript{CE}} we obtain LS distribution shown in Figure~\ref{fig:5}, where all clusters resemble beams heading outwards from the 
center. 

%\unskip

\section{Geometric loss}
\label{GLoss}

In this paper we study a specific form of \textit{L\textsubscript{G}} that defines the positions and radii of clusters. 
Given the cluster center matrix \textit{C} and cluster radius vector 
\textit{r\textsubscript{c}}, \textit{L\textsubscript{G}} is defined as

\begin{equation}
	L_{G} = \sum_{i}^{n_{c}} \sum_{j}^{b_{s}} f_{d}(\sqrt{\sum_{k}^{n_{d}}(z_{jk}(y_{j}=i)-C_{ik})^2},r_{ci}),
	\label{eq:GLoss}
\end{equation}

where \textit{n\textsubscript{c}} is number of classes, \textit{b\textsubscript{s}} is number of samples in training batch, \textit{n\textsubscript{d}} is number of LS 
dimensions, \textit{i} is class index, \textit{j} is input sample index, \textit{k} is LS dimension index, \textit{z\textsubscript{j}} is LS position and 
\textit{y\textsubscript{j}} is true label 
of \textit{j\textsuperscript{th}} sample. \textit{f\textsubscript{d}} is a distance function defined as

\begin{equation}
	f_{d}(x,r_{c}) = exp(ReLU(x - r_{c})) - 1.
	\label{eq:fd}
\end{equation}

\begin{figure}[b]
	\centering
	\includegraphics[scale=0.5]{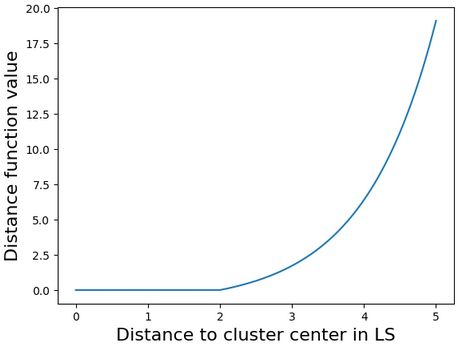} 
	\caption{Curve of \textit{f\textsubscript{d}} function for \textit{r\textsubscript{c}}=2.}
	\label{fig:6}
\end{figure}

Figure~\ref{fig:6} shows \textit{f\textsubscript{d}} curve for \textit{r\textsubscript{c}} = 2 illustrating that this function is constant inside cluster and grows exponentially outside. 
It should be noted that adding 
a constant in (\ref{eq:fd}) is not strictly necessary since derivative of any constant is zero, but we add -1 for convenience so \textit{f\textsubscript{d}} is 
exactly zero inside clusters.  

%\unskip

It is proposed to train SAE with a weighted combination of \textit{L\textsubscript{G}} and \textit{L\textsubscript{CE}} for classification with desired cluster locations. Therefore, combined loss for supervised
training is 

\begin{equation}
	L = L_{CE} + k_{g} \cdot L_{G},
	\label{eq:loss-comb}
\end{equation}

where \textit{k\textsubscript{g}} is geometric loss weight coefficient. In our experiments \textit{k\textsubscript{g}} = 0.2.

\section{Configuring latent spaces for texture classification tasks}
\label{LSconfclassif}

\subsection{Texture classification on small datasets}
\label{smalldata}

In this paper we study LS of SAE designed for texture classification in context of person re-identification \cite{reid-review}. Its purpose is to substitute local binary 
pattern (LBP) \cite{lbpp} in our previously proposed method \cite{combreid}. It should be noted that we do not conduct experiments on re-id benchmarks so re-id accuracy 
metrics are not reported in this study.

Since texture datasets for re-id are not available, a small dataset consisting of real clothes textures from LIP, internet stock photos, and images generated 
using a Stable Diffusion model \cite{focusgit} has been collected. Data is labeled into five classes: uniform (meaning no apparent texture), horizontal lines (hlines), 
vertical lines (vlines), checkered pattern (squares), and dots. We use dataset of as little as 130 images per class and apply a set of augmentations to artificially increase its size. 
The augmentations include rotations of up to 15 deg, perspective change, affine transformation, color jitter, and random erase of 30\% of the image to simulate 
occlusion. Every augmented image is also accompanied by its vertical and horizontal mirror copies. We divide the dataset into 80\% train / 20\% test splits for 
training and testing and consider two testing scenarios: one with train-test sampling done before augmentation (about 130 images in test, no test image augmentation, 
“pre”-split in Table~\ref{tab:2}) and another one with sampling done after augmentation (about 300000 images in test, “post”-split in Table~\ref{tab:2}). In both cases the resulting 
training dataset consists of about 1.2 million train images, which is still quite small for a CV dataset. For generalization study we use 80 images from Market1501 
which are low resolution and poor quality images representing worse case real world scenarios.

\begin{table}
	\caption{Test and generalization accuracy for different random seed runs for SAE trained with different loss functions.} 
	\label{tab:2}
	\centering
	\begin{tabular}{|c|c|c|c|c|c|c|c|c|}
	  \hline
	  \multirow{2}{*}{exp} & \multicolumn{2}{c|}{\textit{L\textsubscript{CE}} pre} & \multicolumn{2}{c|}{\textit{L\textsubscript{CE}} + \textit{L\textsubscript{G}} pre} 
	  & \multicolumn{2}{c|}{\textit{L\textsubscript{CE}} post} & \multicolumn{2}{c|}{\textit{L\textsubscript{CE}} + \textit{L\textsubscript{G}} post} \\ \cline{2-9}
	  
						   & test & gen &	test & gen & test & gen  & test & gen  \\
	  \hline
	  1 & 77 & 62 &	78 &	65 &	81 &	66 &	83	& 67 \\
	  \hline
	  2 & 78 &	68 &	75	& 61	& 82	& 70	& \textbf{84}	& \textbf{71} \\
	  \hline
	  3 & 71 &	62 &	75 &	65 &	81 &	65	& 84	& 61 \\
	  \hline
	\end{tabular}
\end{table}

\begin{figure}[h]
	\centering
	\includegraphics[scale=0.45]{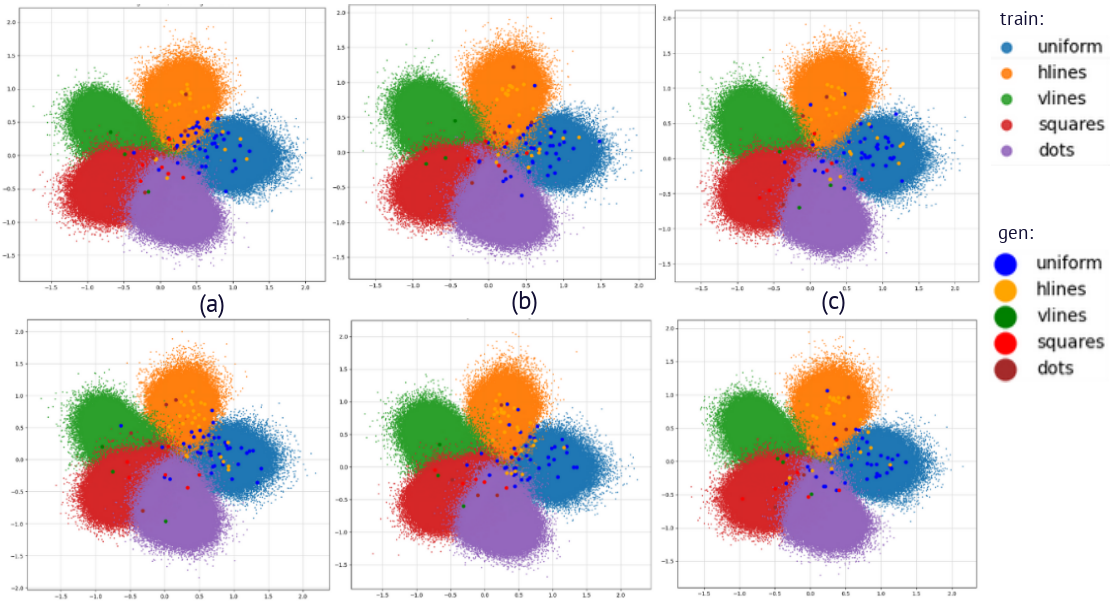} 
	\caption{Positions of Market encodings in LS in \textit{L\textsubscript{CE}} + \textit{L\textsubscript{G}} 
	experiments (a top), 1 pre (a bottom) 1 post, (b top) 2 pre, (b bottom) 2 post, (c top) 3 pre, (c bottom) 3 post.}
	\label{fig:8}
\end{figure}
%\unskip

\begin{figure}[h]
	\centering
	\includegraphics[scale=0.5]{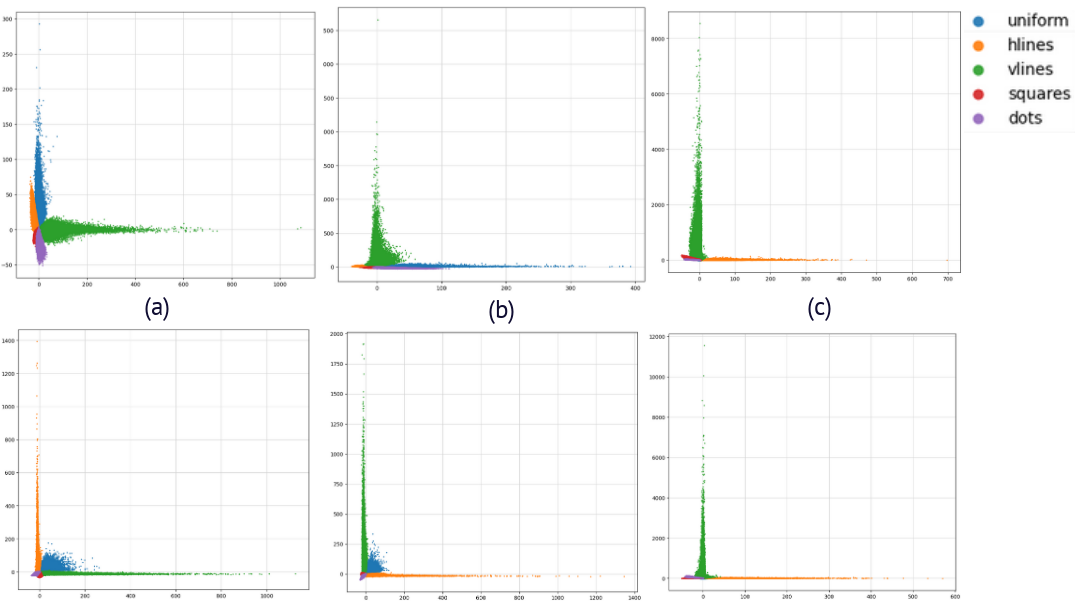} 
	\caption{LS of SAE trained with \textit{L\textsubscript{CE}}, 
	(a top) 1 pre, (a bottom) 1 post, (b top) 2 pre, (b bottom) 2 post, (c top) 3 pre, (c bottom) 3 post.}
	\label{fig:7}
\end{figure}
%\unskip

\subsection{Conventional training and LS configuration training}
\label{CEvsCEG}

Table~\ref{tab:2} shows experimental results obtained by performing three runs with different random seeds fixed for four experiments constituting the run. 
Random variables include SAE initialization weights, train/test split samples, augmentation color jitter parameters, batch train data splits. All models are trained 
for 50 epochs with 1e-6 learning rate, with training accuracy exceeding 99\% for all experiments. The classification results shown in Table~\ref{tab:2} illustrate that 
test and Market1501 generalization accuracies are higher for the proposed training method. Using “post”-split in pre-processing also leads to better results. 

\begin{figure}%[h]
	\centering
	\includegraphics[scale=0.4]{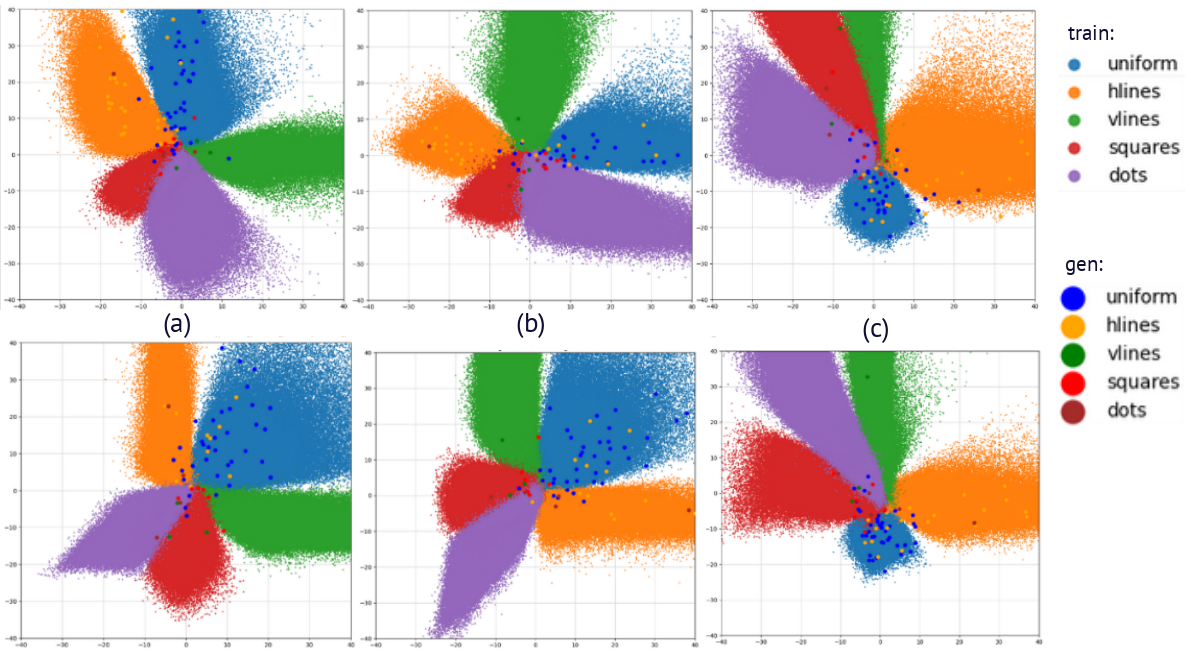} 
	\caption{Positions of Market encodings in LS in \textit{L\textsubscript{CE}} experiments 
	(a top), 1 pre (a bottom) 1 post, (b top) 2 pre, (b bottom) 2 post, (c top) 3 pre, (c bottom) 3 post.}
	\label{fig:9}
\end{figure}

Geometric loss configurations has the following parameters: five clusters \textit{n\textsubscript{c}} = 5 that are at \textit{d\textsubscript{c}} = 0.85 distance from 
center (0,0) and $2\pi/\textit{n\textsubscript{c}}$ = 
72\textdegree \ angles relative to the 
neighbors with cluster radius \textit{r\textsubscript{c}} = 0.34. 
Figures~\ref{fig:8} - \ref{fig:9} show positions of LS embeddings of train data (small points) with Figures~\ref{fig:7} and \ref{fig:9} also showing Market 
generalization data LS 
embeddings (large points). Figure~\ref{fig:8} (a) - (c) shows that \textit{L\textsubscript{CE}} + \textit{L\textsubscript{G}} training method allows to achieve the 
desired LS distributions for all experiments. Clusters resemble petals due to large number of points and the interplay between 
\textit{L\textsubscript{CE}} and \textit{L\textsubscript{G}}. 
Figure~\ref{fig:8} shows that 
in all cases clusters appear in the desired positions with all training points being roughly inside a circle with radius two. All LS encodings in the Market transfer 
experiments also lie withing the specified regions not leaving the configured LS. This illustrates that even that LS condition is present only during training, the 
learned weights succeed in projecting new unseen data into the specified areas in LS. 

%\unskip

On the contrary, for conventional training \textit{L\textsubscript{CE}} allows \textit{z} to take any value and in this case, we can see huge “explosion” in LS positions along axes in Figure~\ref{fig:7}. 
Moreover, the LS shown in Figure~\ref{fig:7} (a) - (f) are totally different. It is further illustrated by Figure~\ref{fig:9} where [-40,40] regions are visualized. 
There is a 
high degree of asymmetry between clusters, and clusterization patterns are very different for different sets of random starting conditions. The comparison of 
Figures~\ref{fig:8} and \ref{fig:9} shows positive effects of \textit{L\textsubscript{G}} on training making it more stable and predictable.

\section{Image similarity estimation using LS analysis}
\label{LSsim}
\subsection{Similarity in LS}
\label{LSsimtheory}

Having information about cluster location and size that pre-configured LS provides can be an advantage for similarity estimation tasks. Since LS projections are 
confined into a compact space, we can estimate how similar the inputs are just by analyzing their positions and distances between their LS encodings. Conventional 
LS does not have this property since clusters have different shapes and their sizes are not predetermined, as Figures~\ref{fig:7} and \ref{fig:9} illustrate.

To evaluate the similarity between two inputs we first propose to evaluate the similarity of each input to all classes by analyzing their LS positions relative to 
cluster centers. This allows to obtain class similarity vectors and then use these vectors to compare the inputs. To achieve that, vectors of cluster center distances 
\textit{d\textsubscript{i}} are calculated as Euclidean distances to cluster centers as 

\begin{equation}
	d_{ji} = \sqrt{\sum_{k}(z_{jk} - C_{ik})^2}.
	\label{eq:dji}
\end{equation}

They are used to calculate class similarity vectors as

\begin{equation}
	v_{ji} = 
	\begin{cases}
	1 - k_{b} \cdot sin(\frac{d_{ji}} {r_{ci}})\ if\ d_{ji} \le r_{ci}, \\
	ReLU(\frac{b_{c} \cdot (R_{d} - d_{ji})} {R_{d} - r_{ci}})\ otherwise; \\
	\end{cases}
	\label{eq:vji}
\end{equation}

where \textit{b\textsubscript{c}} is a parameter corresponding to the desired similarity value at cluster boundary when \textit{d\textsubscript{ji}} 
= \textit{r\textsubscript{ji}}; \textit{k\textsubscript{b}} is calculated as

\begin{equation}
	k_{b} = \frac{1-b_{c}}{sin(1)},
	\label{eq:kb}
\end{equation}

and is a constant for given \textit{b\textsubscript{c}}, and \textit{R\textsubscript{d}} is the distance between neighboring clusters

\begin{equation}
	R_{d} = d_{c} \ \sqrt[]{2\cdot(1 - cos(\frac{2\pi}{n_{c}}))}.
	\label{eq:Rd}
\end{equation}

Cluster parameters in Section~\ref{CEvsCEG} yeild \textit{R\textsubscript{d}} = 1, and \textit{b\textsubscript{c}} = 0.79 in our experiments. 
Finally, for two points their similarity is calculated as

\begin{equation}
	sim_{12} = \frac{\sum_{i} min(v_{1i},v_{2i})}{max(sum(v_{1}),sum(v_{2}))}.
	\label{eq:sim}
\end{equation}

\subsection{LS similarity for texture retrieval}
\label{LSsimfortex}

\begin{figure}[b]
	\centering
	\includegraphics[scale=0.30]{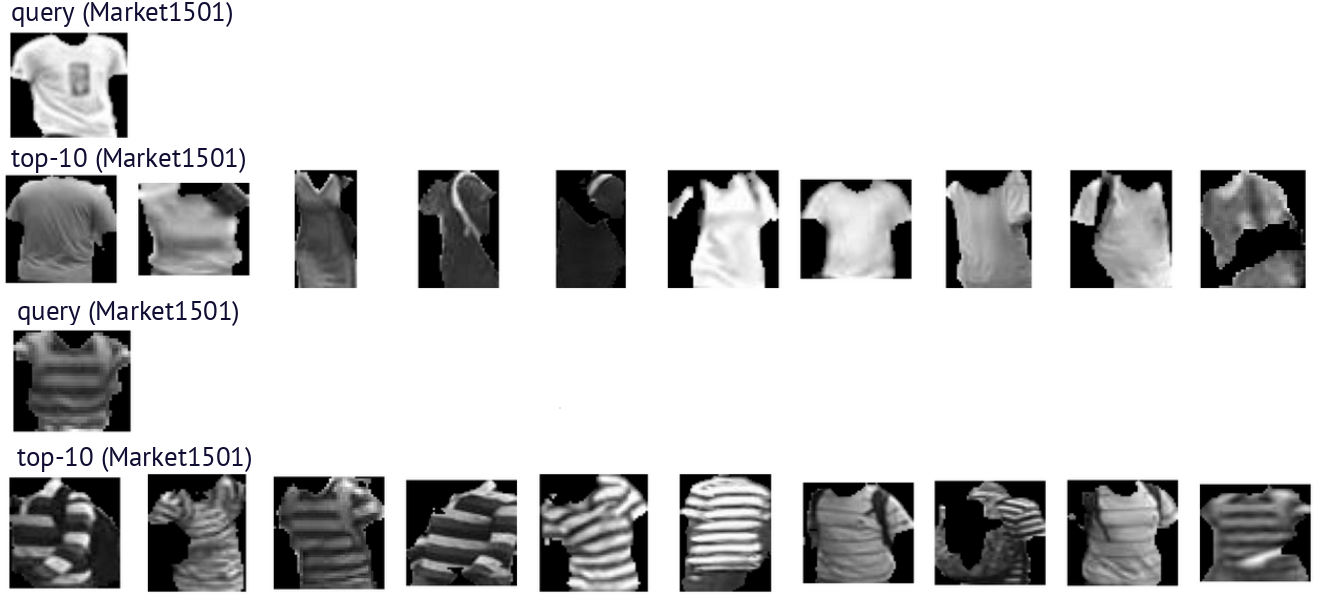} 
	\caption{Examples of top-10 images matching texture of query image in Market1501 dataset.}
	\label{fig:11}
\end{figure}
%\unskip

Using the proposed similarity measure we can make similarity ranking of textures in LIP, Market1501, and Wildtrack datasets. It should be noted that LIP is a 
human parsing dataset, whereas Market1501 and WildTrack are re-id datasets. None of the datasets are specifically labeled for texture classification. Moreover, 
whereas subsets of LIP and Market1501 have been labeled as part of this study, WildTrack have not been labeled, which does not prevent us from creating similarity ranking 
thus surpassing the limitations of conventional classification methods. That is, whereas five classes have been used to train SAE, other class data can be analyzed 
with some degree of accuracy by estimating its similarity to the train classes. 

It should be noted that only LS encodings are needed to conduct dataset search with the proposed method. Therefore, only one encoder forward pass is needed for each 
image in database, and LS encodings can be saved in the database along with images. Then, after query image is encoded and its LS encoding is known, it can be compared 
with database LS encodings directly. This is very fast and efficient since it reduces the involvement of NN in the analysis, and since 
ranking computations can easily be performed on general purpose CPU, this reduces the overall hardware requirements of the method. 

Figure~\ref{fig:11} shows top-10 clothes with textures most similar to the texture in query image (see Appendix~\ref{app} for more examples). It should be stressed that all results are obtained 
with SAE trained on our dataset and no retraining for specific datasets was conducted. This illustrates good generalization of the 
method to previously unseen data of different quality and type.

\begin{figure}%[h]
	\centering
	\includegraphics[scale=0.35]{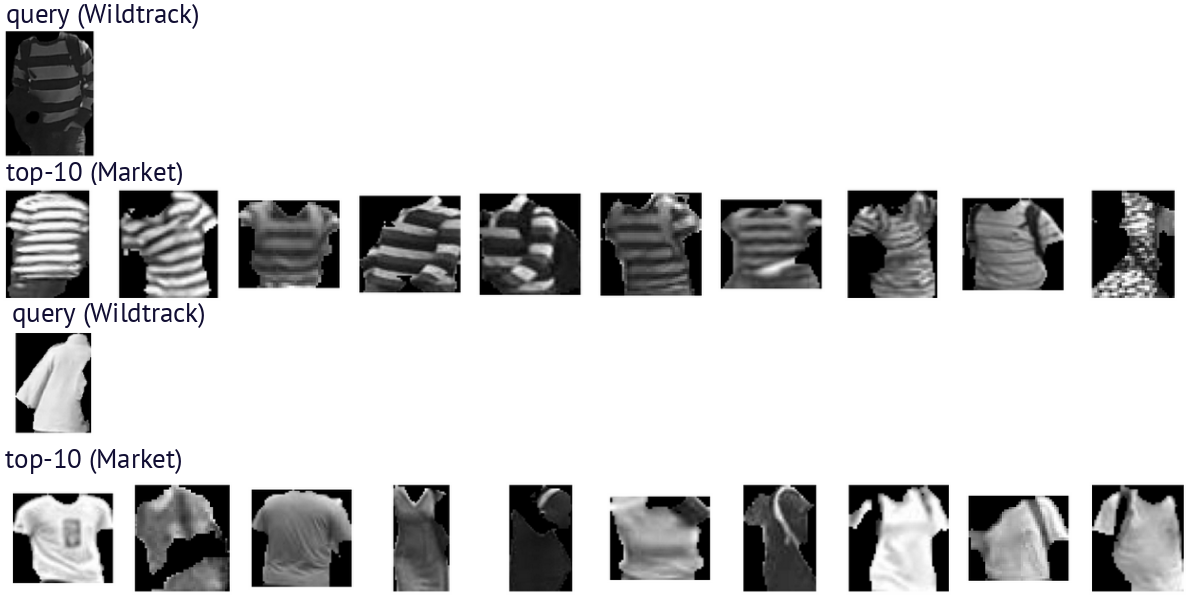} 
	\caption{ Top-10 similar images for Wildtrack query searched in Market1501.}
	\label{fig:13}
\end{figure}

Figure~\ref{fig:13} shows cross-dataset texture retrieval results where a similarity ranking for a query image from one dataset searched in another dataset is 
created (more in Appendix~\ref{app}). Specifically, queries from Market are searched in Wildtrack and vice versa using a model trained on data containing images from 
neither dataset. These 
examples show high visual similarity between the query and ten most similar images further illustrating great generalization capabilities of the proposed method. 
That is, images from very different cameras have been used for training and searching meaning that the model is trained to perform accurately over wide range of 
different lighting conditions and variations in image quality with no fine-tuning.

%\unskip

\subsection{Dataset search without a query image}
\label{Noq}

\begin{figure}%[h]
	\centering
	\includegraphics[scale=0.45]{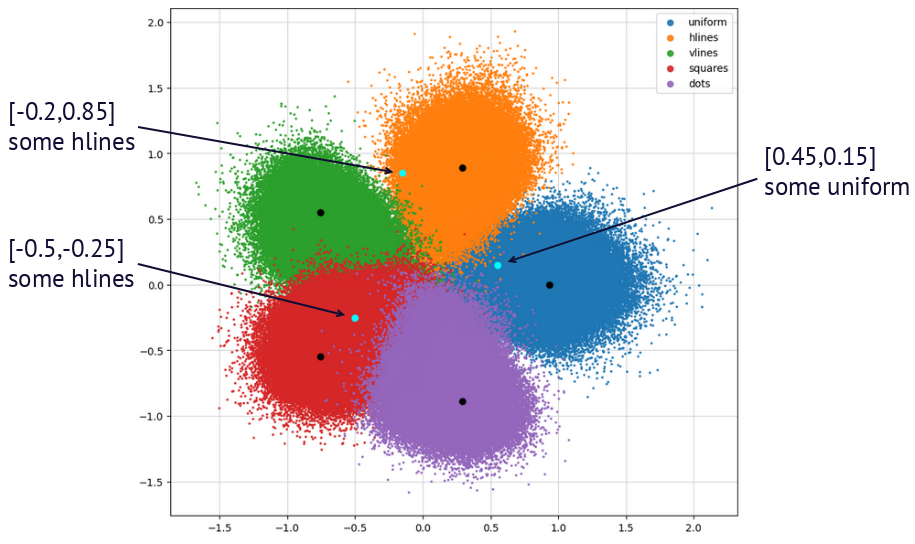} 
	\caption{Example correlations between text queries and LS positions used for similarity search. Black dots correspond to configuration centers.}
	\label{fig:15}
\end{figure}

We previously proposed an image retrieval method that allowed us to conduct database search using color description without query image \cite{combreid}. It was done by 
creating feature vectors for typical colors and using such vectors in a search algorithm by comparing them to feature vectors of images stored in database. In this 
paper we show how the same principle can be used for texture search. Figure~\ref{fig:15} shows how preconfigured LS can be used to correlate text queries with LS regions. 
Whereas it is possible to simply use cluster center locations as encodings, we can search for less clear examples to account for influence of specific cameras and 
lighting conditions by choosing a random point from relevant cluster. 
Furthermore, LS encodings can be estimated for texture types that are not in training classes as some combination of training class features. 

%\unskip

Figure~\ref{fig:16} shows the retrieval results for points shown in Figure~\ref{fig:15} (more examples in Appendix~\ref{app}). 
Uniform textures are retrieved with samples showing deviations 
from perfect uniform texture like folds and prints. In Figure~\ref{fig:17} horizontal line pattern textures are successfully retrieved from Market1501 
apart from one sample with unclear 
texture due to low resolution. In Figure~\ref{fig:18} checked pattern coat has the highest score along with unspecified “rough” texture coats in Widltrack. 
It should be noted that due to 
low number of images in Wildtrack subset used for search there are simply very few images similar to the query. 
Nevertheless, the ability to access similarity with unlabeled unspecified textures is a great advantage of the proposed method.

\begin{figure}%[h]
	\centering
	\includegraphics[scale=0.45]{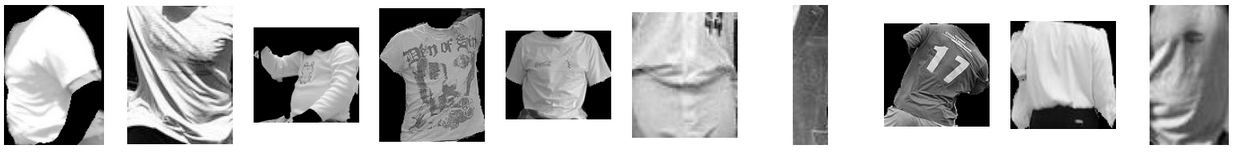} 
	\caption{Search results for some uniform texture in training dataset.}
	\label{fig:16}
\end{figure}
%\unskip

\section{Discussions}
\label{discussions}

\subsection{LS topology}
\label{LStopology}

In previous Sections we have shown that it is possible to add a relatively simple term to loss function to achieve a predefined configuration of LS during training. 
This allows to define location and size of the clusters which is then used to define an LS similarity measure to evaluate the similarity of different inputs based on 
their LS projections. Furthermore, Figure~\ref{fig:8} shows that different training runs result in exactly the same LS train distribution for \textit{L\textsubscript{CE}} 
+ \textit{L\textsubscript{G}}, whereas 
conventional training leads to strikingly different LS clusterization. Therefore, adding \textit{L\textsubscript{G}} term makes training more stable and predictable which in turn 
improves the interpretability of the results.

While in this paper the \textit{L\textsubscript{G}} was formulated for two-dimensional case for purposes of LS visualization on a 2D plane, it can be easily extended to n-dimensional cases. 
However, the possibility to directly affect the LS topology can have more profound implications. It is possible to obtain more “exotic” topologies with interesting 
properties through encoder configuration by guiding LS projection terms with special equations similar to (\ref{eq:pol-r})-(\ref{eq:z_new}). While this has been shown for polar coordinates, 
it is possible to obtain non-Euclidian spaces, for instance, hyperbolic spaces where distances between clusters are calculated in a completely different manner. 
The usefulness of this is not obvious and has to be studied in more detail. 

\subsection{Improved generalization with geometric loss}
\label{GGen}

Sections~\ref{LSconfclassif} and \ref{LSsim} have shown better generalization of SAE trained with \textit{L\textsubscript{G}}. This can be partly explained by 
more structured and compact LS. 
It should be mentioned that test accuracy is not very high for both conventional and proposed training methods because of the small dataset size.
That is, there is a possibility that test set contains samples significantly different from the train set simply due to the high variance among data samples. 
There also are significant challenges for texture classification due to the inherent similarity between certain classes. The analysis of the incorrect classifications 
has shown that folds on uniform texture often get misclassified as vertical or horizontal line patterns, and sometimes patterns of squares and dots might become 
indistinguishable due to low resolution of input images (see Table~\ref{tab:2}). This can be attributed to the nature of the experiment used in this study and not to the 
proposed LS configuration methodology. 

High generalization of the method is also illustrated by Figure~\ref{fig:13} where images from one dataset are used as queries for search in another dataset. 
It should be stressed again that for both experiments SAE was trained on third dataset, indicating that the proposed method can work in real-world scenarios where 
neither database images nor new images obtained from cameras can be used for training. The proposed geometric clusterization in combination with the proposed similarity 
estimation method has an emerging property of the ability to access similarity for classes that are not in training class set. When given a completely new texture, 
SAE encodes it in LS as any other input, and its position depends on its similarity to train classes via proximity to class clusters. Conventional SAE classifier then 
outputs a label which is contained in training classes’ set thus being unable to predict an unseen class. However, given a pair of unseen class inputs the proposed 
method can access their similarity directly from their LS positions, giving high similarity score to similar inputs. 
This is a significant advantage of the proposed method when input similarity estimation rather than classification is to be performed.

\subsection{Text query search without a language model}
\label{discq}

In Section~\ref{Noq} we have shown that we can estimate regions of LS where certain types of textures are likely to be. This is possible since we have configured 
LS in a predefined manner. This allows us to establish correspondence between verbal or text descriptions of textures and LS regions. The results in 
Figure~\ref{fig:16}, \ref{fig:17}, and \ref{fig:18} that show the successful text query search for different datasets illustrate good generalization capabilities. 
Hence, a necessity to process natural language with a separate model to create text embeddings is removed. This idea can be extended to generative models, too, 
in order to reduce the number of components and improve interpretability. However, this would require training a diffusion model with \textit{L\textsubscript{G}} and 
preconfiguring 
LS with \textit{n\textsubscript{d}} much larger than 2, e.g. 4096 for conventional VAEs. The means of achieving this are not obvious and this topic will be studied in the future.

\section{Conclusions}
\label{conclusions}

This paper proposes two methods for LS configuration for SAE. The first method implements geometric loss term which guides training to create the desired 
configuration of the LS. The second method modifies SAE encoder to alter LS properties. It has the potential to provide unprecedented control over LS properties 
and guarantees the desired behavior during inference. However, it causes difficulties during training which will be studied in the future. For the former it is shown 
that training with the proposed geometric loss reliably leads to the desired LS configuration while making training more stable and interpretable. It also leads to 
higher generalization accuracy, as shown by clothes texture classification task results for SAE trained on a small custom dataset and tested on subsets from datasets LIP, 
Market1501, and WildTrack. Using LS configuration also allows to define a similarity measure directly in LS to access similarity of multiple inputs without 
classifier. This is used to create similarity ranking and search for images with textures similar to a given query image. High generalization of the proposed method 
is illustrated by successful inter- and cross-dataset searches where a model trained on one dataset is used to search for a query from second dataset in third dataset. 
Finally, known LS configuration is used to correspond text queries to LS positions and conduct database search without query images and language models.

\section*{Acknowledgement}
\label{acknowledgement}

The author would like to thank Dr Igor Netay and Dr Anton Raskovalov for fruitful discussions, and Vasily Dolmatov for his assistance in problem formulation, 
choice of methodology, and supervision.

\section*{Data availability}
\label{Data-Statement}

Training data used in this study can be provided on request addressed to the corresponding author.

%\section*{References}
%\label{references}

\bibliographystyle{IEEEtran}
\bibliography{IEEEabrv,ms}

\newpage
\appendix
\renewcommand\thefigure{\thesection.\arabic{figure}}    
\section{Appendix 1}
\label{app}
\setcounter{figure}{0}  

\begin{figure}[h]
	\centering
	\includegraphics[scale=0.35]{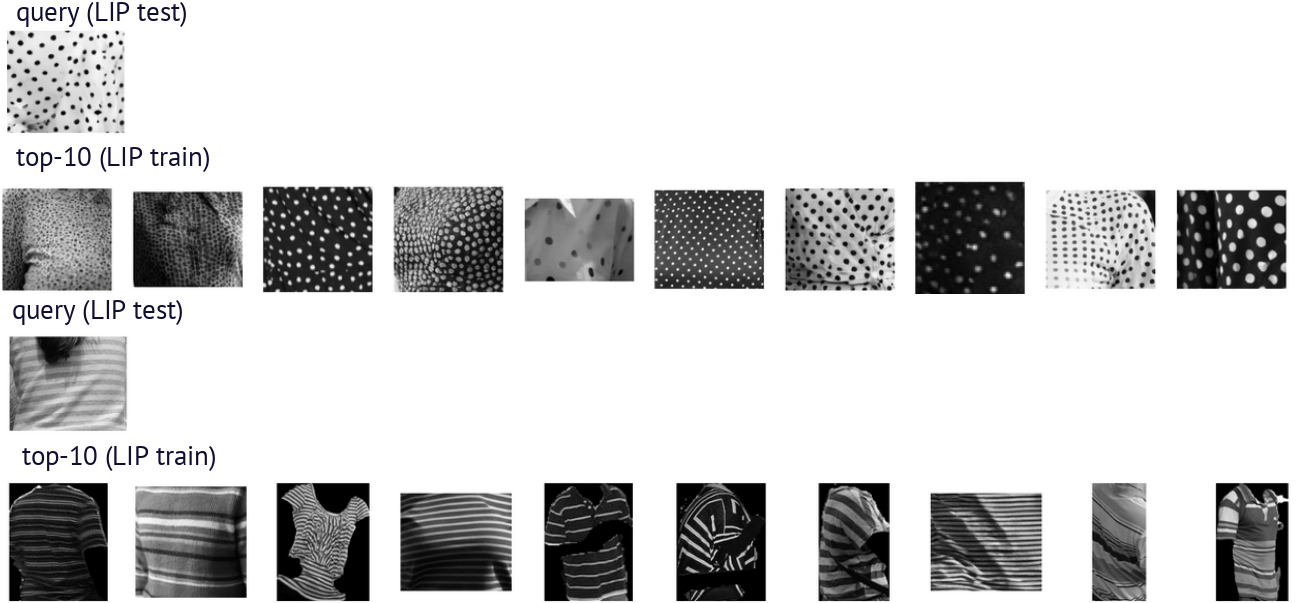} 
	\caption{Examples of top-10 images matching texture of query image from LIP test searched in LIP train subsets.}
	\label{fig:10}
\end{figure}
%\unskip

\begin{figure}[h]
	\centering
	\includegraphics[scale=0.35]{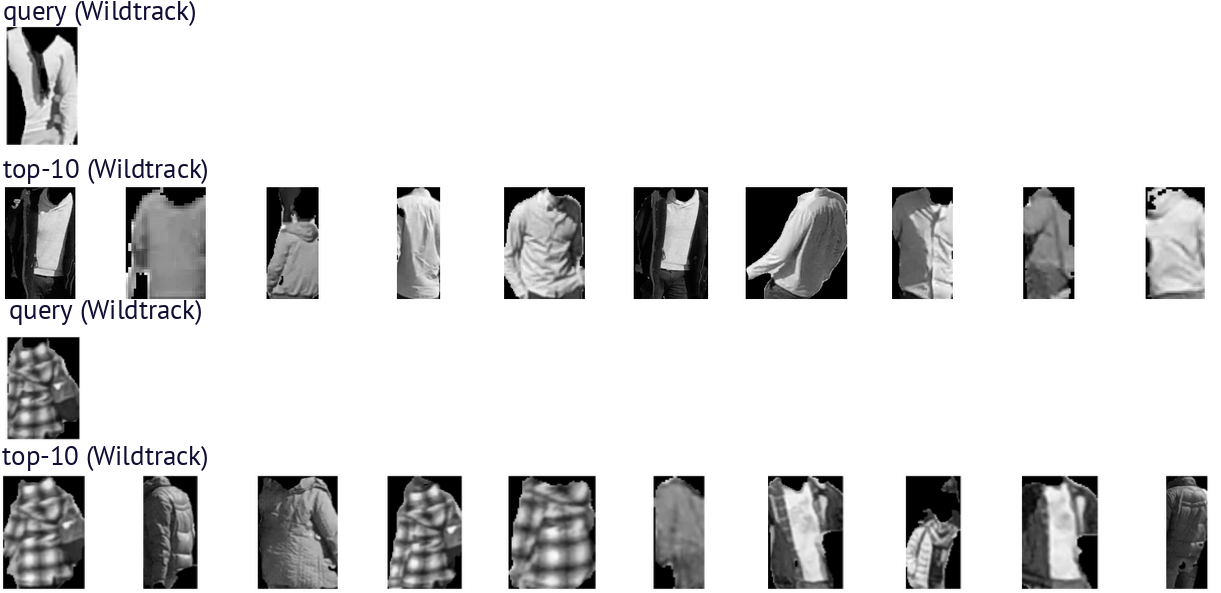} 
	\caption{Examples of top-10 images matching texture of query image on Wildtrack dataset.}
	\label{fig:12}
\end{figure}
%\unskip

\begin{figure}%[h]
	\centering
	\includegraphics[scale=0.35]{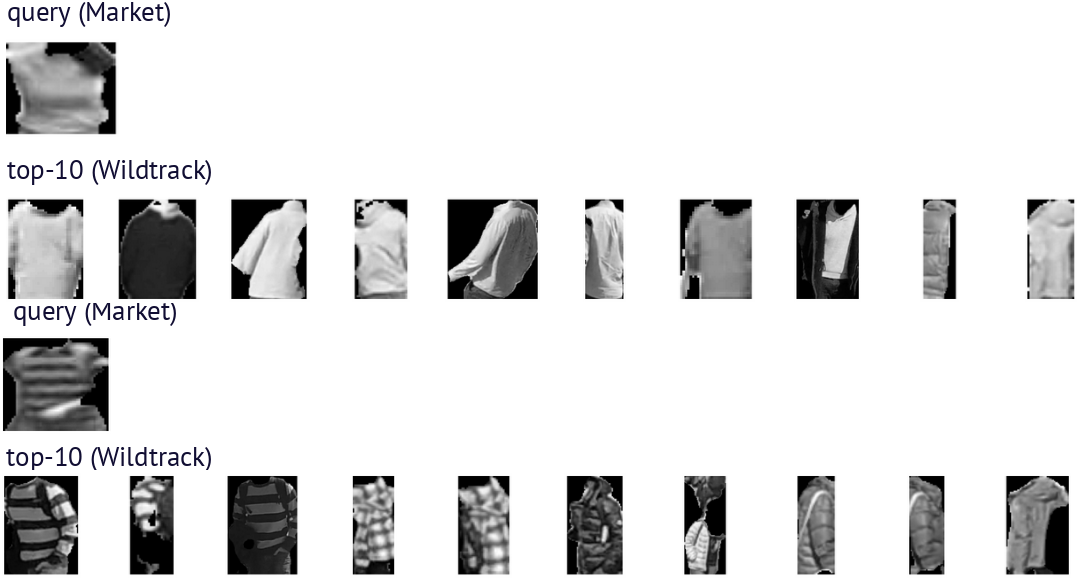} 
	\caption{Top-10 similar images for Marhet1501 query searched in Wildtrack.}
	\label{fig:14}
\end{figure}
%\unskip

\begin{figure}%[h]
	\centering
	\includegraphics[scale=0.45]{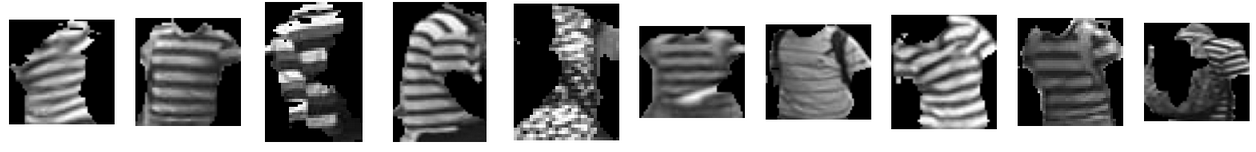} 
	\caption{Search results for some horizontal lines texture in Market1501 dataset.}
	\label{fig:17}
\end{figure}
%\unskip

\begin{figure}%[h]
	\centering
	\includegraphics[scale=0.45]{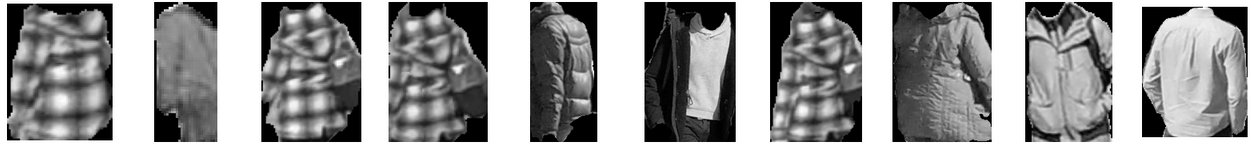} 
	\caption{Search results for textures with checkered pattern in Wildtrack dataset.}
	\label{fig:18}
\end{figure}
%\unskip

\end{document}